\documentclass[11pt,a4paper]{article}
\usepackage[hyperref]{emnlp2020}
\usepackage{times}
\usepackage{amsmath}
\usepackage{booktabs}
\usepackage{makecell}
\usepackage{multirow}
\usepackage{pifont}
\usepackage{multicol}
\usepackage{amssymb}
\usepackage{latexsym}

\usepackage{microtype}
\usepackage{bbold}
\usepackage{amsmath}
\usepackage{ulem}
\usepackage{graphicx}
\usepackage{amsmath}

\usepackage{booktabs}

\usepackage{footmisc} 
\usepackage{multirow}
\aclfinalcopy % Uncomment this line for the final submission
\usepackage{hyperref}

\makeatletter
\newcommand{\printfnsymbol}[1]{%
  \textsuperscript{\@fnsymbol{#1}}%
}

\title{Why Skip If You Can Combine: A Simple Knowledge Distillation Technique for Intermediate Layers}

\author{Yimeng Wu\Thanks{ These authors contributed equally.}\\
        \And
        Peyman Passban\printfnsymbol{1}\\
        \hspace{40mm} Huawei Noah’s Ark Lab\\
        \hspace{41mm}\texttt{firstname.lastname@huawei.com} \\
        \And
        Mehdi Rezagholizadeh\\
        \And
        Qun Liu\\
        }

\date{}

\begin{document}
\maketitle
\begin{abstract}
With the growth of computing power neural machine translation (NMT) models also grow accordingly and become better. However, they also become harder to deploy on edge devices due to memory constraints. To cope with this problem, a common practice is to distill knowledge from a large and accurately-trained teacher network ($\mathcal{T}$) into a compact student network ($\mathcal{S}$). Although knowledge distillation (KD) is useful in most cases, our study shows that existing KD techniques might not be suitable enough for deep NMT engines, so we propose a novel alternative. In our model, besides matching $\mathcal{T}$ and $\mathcal{S}$ predictions we have a combinatorial mechanism to inject layer-level supervision from $\mathcal{T}$ to $\mathcal{S}$. In this paper, we target low-resource settings and evaluate our translation engines for Portuguese$\rightarrow$English, Turkish$\rightarrow$English, and English$\rightarrow$German directions. Students trained using our technique have $50\%$ fewer parameters and can still deliver comparable results to those of $12$-layer teachers. 
\end{abstract}

\section{Introduction}
Almost in all deep learning tasks, including neural machine translation (NMT), an ensemble of models outperforms a single model. In fact, ensemble modelling (training multiple models and ensemble decoding) is supported by most publicly available NMT frameworks \citep{opennmt,mariannmt,tensor2tensor,ott2019fairseq}. However, we know that dealing with multiple models could be challenging, especially in deep learning scenarios. To tackle the issue, one effective solution is to compress the knowledge in an ensemble into a single model through distillation \citep{bucilua2006model,hintonkd}.

The core part of any knowledge distillation (KD) pipeline is a component that matches different models' predictions, which is usually implemented via multiple cost functions (see Section \ref{background}). Furthermore, we also need to take care of the architecture mismatch that may exist between student ($\mathcal{S}$) and teacher ($\mathcal{T}$) models. In KD, these two models can have different architectures \citep{jiao2019tinybert,pkd} and the motivation is to be able to compress a large teacher into a smaller student.   

This research focuses on the aforementioned issue. If we distill from intermediate layers of a teacher that has more layers than its student, we have to select a subset of $\mathcal{T}$ layers and skip others as there are no peers for all of them on the $\mathcal{S}$ side. Clearly, we do not benefit from the skipped layers in this scenario. This type of KD introduces a problem of finding an optimal subset of $\mathcal{T}$ layers (to distill from). Although this might, to some extent, be mitigated via a search mechanism, our experimental results show that the problem is severe in NMT and each layer plays a unique role. Therefore, we prefer to keep all layers rather than skip them. 

KD has recently become popular in NMT but,  to the best of our knowledge,  all NMT models \citep{kim2016sequence,tan2019multilingual} are still trained using the original idea of KD \citep{hintonkd}, which is referred to as Regular KD (\textit{RKD}) throughout this paper. \textit{RKD} only matches $\mathcal{S}$ and $\mathcal{T}$ outputs, regardless of their internal architecture. However, there exist techniques such as Patient KD (\textit{PKD}) \citep{pkd} proposed for other tasks that not only match final predictions but also focus on internal components and distill their information too \citep{sun2020mobilebert}. In this research, we borrowed those ideas and adapted them to NMT. This is the first contribution of the paper.

\textit{PKD} and other similar models suffer from the \textit{skip} problem, which happens when $\mathcal{T}$ has more layers than $\mathcal{S}$ and some $\mathcal{T}$ layers have to be skipped in order to carry out layer-to-layer distillation. In this paper, we propose a model to distill from \textit{all} teacher layers so we do not have to skip any of them. This is our second contribution by which we are able to outperform \textit{PKD}. Moreover, for the first time we report experimental results for Transformer-based \citep{vaswani2017attention} models trained with a layer-level KD technique in the context of NMT. This set of results is our third and last contribution in this paper. 

The remainder of the paper is organized as follows: In Section \ref{background} we explain the fundamentals of KD. Section \ref{method} discusses the methodology. We describe the advantages of our model and accompany our claims with experimental results in Section \ref{exp}. Finally, in Section \ref{future}, we conclude the paper with our future plan. 

\section{Background}\label{background}
Usually, in multi-class classification scenarios the training criterion is to minimize the negative log-likelihood of samples, as shown in Equation \ref{eq:0}:
\begin{equation}\label{eq:0}
\mathcal{L}(\theta) = - \sum_{v=1}^{|V|} \mathbb{1}(y=v) \times \log p(y=v|x;\theta)  
\end{equation}
where $\mathbb{1}$(.) is an indicator function, $(x,y)$ is an input-output training tuple, and $\theta$ and $|V|$ are the parameter set of the model and the number of classes, respectively. There is no feedback returned from the network for misclassified examples as $\mathbb{1}(y\neq v)=0$. This issue is resolved in KD with extending $\mathcal{L}$ with an additive term \citep{kim2016sequence,tan2019multilingual}, as shown in Equation \ref{eq:2}:
\begin{multline}\label{eq:2}
    \mathcal{L}_{KD}(\theta_{\mathcal{T}},\theta_{\mathcal{S}}) =\\ - \sum_{v=1}^{|V|} q(y=v|x;\theta_{\mathcal{T}})  \times \log p(y=v|x;\theta_{\mathcal{S}})
\end{multline}
where there is a student model with the parameter set $\theta_S$ whose predictions are penalized with its own loss as well as $\mathcal{T}$ predictions given by $q(y=v|x;\theta_{T})$. In KD, the first component of the loss ($q$) is usually referred to as the \textit{soft} loss and the $\mathcal{S}$ model's loss is known as the \textit{hard} loss. This form of training provides richer feedback compared to the previous one and leads to high(er)-quality results. KD for NMT also follows the same principle where $V$ is a target-language vocabulary set and $\mathcal{L}_{KD}$ is computed for \textit{each} word during decoding. 

With the matching strategy proposed in KD, $\mathcal{S}$ learns to mimic its $\mathcal{T}$. A teacher could be a deep model trained on a large dataset but we do not necessarily need to have the same complex architecture for $\mathcal{S}$. We can distill teacher's knowledge into a smaller model and replicate its results with fewer resources.

\citet{kim2016sequence} studied this problem and proposed a sequence-level extension of Equation \ref{eq:2} for NMT models. They evaluated their idea on recurrent, LSTM-based models \citep{hochreiter1997long} and could run the final model on a cellphone. %In this paper we tried to report the impact of their idea in Transformers. 
\citet{freitag2017ensemble} extended the original two-class idea (one $\mathcal{S}$ with one $\mathcal{T}$) to distill from multiple teachers. They trained an attention-based recurrent model \citep{bahdanau2015neural} for their experiments. 

\citet{tan2019multilingual} proposed a setting to train a multilingual Transformer for different language directions. In order to have a high-quality multilingual model they distill knowledge from separately trained bilingual models. Their work is one of the few papers that reports KD results for NMT on Transformers. However, their results are not directly comparable to ours as they benefit from rich, multilingual corpora. 

\citet{wei2019online} introduced a pipeline where a student model learns from different checkpoints. At each validation step, if the current checkpoint is a better model than the best existing checkpoint, $\mathcal{S}$ learns from it, otherwise the best stored checkpoint is considered as the teacher.

In all models discussed so far, \textcolor{blue}{\textit{i}}) $\mathcal{S}$ usually has the same architecture as its teacher(s) but we know that recent NMT models, particularly Transformers, are deep models which makes them challenging to run on edge devices. Moreover, \textcolor{blue}{\textit{ii}}) the training criterion in the aforementioned models is to combine final predictions. Transformers have new components (e.g. self-attention) and multiple (sub-)layers that consist of valuable information \citep{clark2019does} and we need more than an output-level combination to efficiently distill for/from these models. Therefore, a new technique that is capable of addressing \textcolor{blue}{\textit{i}} and \textcolor{blue}{\textit{ii}} is required.

Authors of \textit{PKD} spotted the problem and focused on internal layers \citep{pkd}. They studied the limitations of \textit{RKD} for BERT \citep{devlin2019bert} models and introduced a layer-to-layer cost function. They select a subset of layers from $\mathcal{T}$ whose values are compared to $\mathcal{S}$ layers. They also showed that different internal components are important and play critical roles in KD. 

The layer-level supervision idea was successful for monolingual models but so far, no one has tried it in the context of NMT. In this paper, we investigate if the same idea holds for bilingual models or if NMT requires a different type of KD. Moreover, we address the \textit{skip} problem in \textit{PKD} (shown in Figure~\ref{fig:1}). It seems in deep teacher models we do not need to skip layers and we can distill from \textit{all layers}.

\section{Methodology}\label{method}
\begin{figure}[h]
\begin{center}
\includegraphics[scale=0.34]{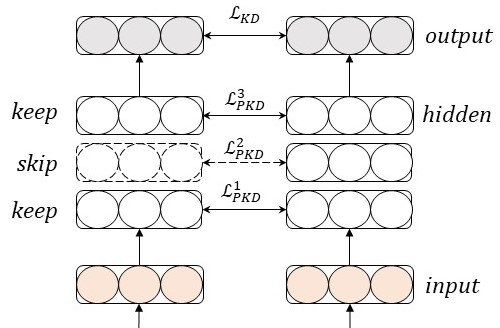}
\end{center}
\caption{\label{fig:1} The network on the left-hand side is $\mathcal{S}$ and the other one is $\mathcal{T}$. In this example, $\mathcal{T}$ has $3$ hidden layers and KD for intermediate layers can be applied using all layers or a subset of them, e.g. the second layer can be skipped.}
\end{figure}

In \textit{RKD}, distillation only happens at the output level whereas \textit{PKD} introduces layer-wise supervision. This idea is illustrated in Figure \ref{fig:1}.  

In \textit{PKD}, finding a skippable layer is the main challenge. Accordingly, we propose a combinatorial idea, \textit{CKD}, by which we are able to fuse layers and benefit from \textit{all} information stored in \textit{all} layers. Our idea can be formulated as follows: \begin{equation}\label{eq:3}
\begin{split}
    \mathcal{L}_{CKD}(L_s,L_t) =& \sum_{l^i_s \in L_s} MSE(l^i_s,f_t^i) \\
    f^i_t =& F(l^j_t); j \in M(i)
\end{split}
\end{equation}
where $L_s$ and $L_t$ indicate the set of all hidden layers of $\mathcal{S}$ and $\mathcal{T}$, respectively. $MSE()$ is the mean-square error function and $l_s^i$ is the $i$-th hidden layer of $\mathcal{S}$. In \textit{PKD}, $f_t^i$ is the teacher's $i$-th layer whereas in our case $f_t^i$ is the result of a fusion applied through the function $F()$ to a particular subset of $\mathcal{T}$ layers. This subset is defined via a mapper function $M()$ which takes an index (pointing to a layer on the student side) and returns a set of indices from the teacher model. Based on these indices, teacher layers are combined and passed to the distillation process, e.g. if $M(2)=\{1,3\}$ that means $F$ is fed by the first ($l^1_t$) and third ($l^3_t$) layers of $\mathcal{T}$ and the distillation happens between $l^2_s$ and $f_t^2$ (result of fusion).

For $F()$, a simple concatenation followed by a linear projection provided the best results in our experiments, so in the previous example:
$$f_t^2 = F(l^1_t,l^3_t) = W[l^1_t\bullet l^3_t]^T + b$$
where $\bullet$ indicates concatenation, and $W\in\mathbb{R}^{d \times 2d}$ and $b \in\mathbb{R}^d$ are learnable parameters of KD. All $l^1_t$, $l^3_t$, $l^2_s$, and $f_t^2$ are $d$-dimensional vectors. 

\begin{figure}[h]
\begin{center}
\includegraphics[scale=0.3]{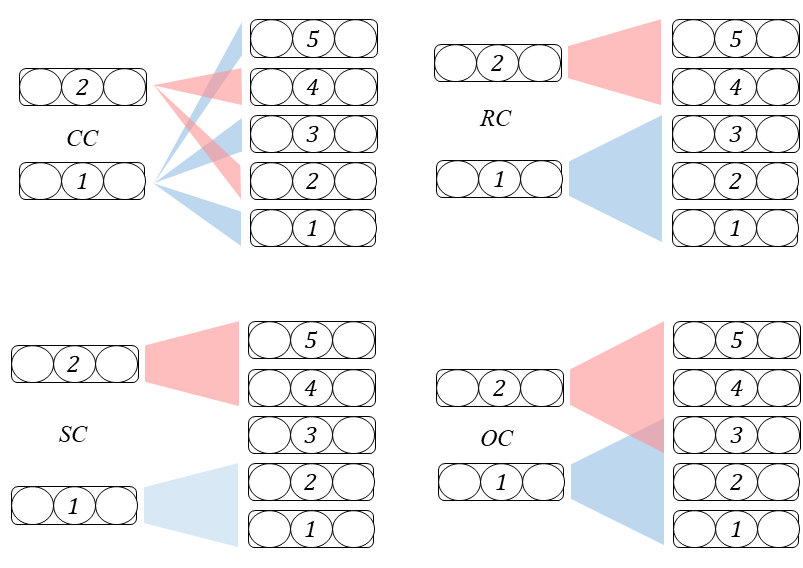}
\end{center}
\caption{\label{fig:2} Different variations of \textit{CKD}. $\mathcal{T}$ has $5$ and $\mathcal{S}$ has $2$ hidden layers. For the \textit{CC} case $M(1)=\{1,3,5\}$.}
\end{figure}

The mapper function $M()$ defines our combination strategy for which we have $4$ different variations of regular combination \textit{(RC)}, overlap combination \textit{(OC)}, skip combination \textit{(SC)}, and cross combination \textit{(CC)}. Figure \ref{fig:2} visualizes these variations. As the figure shows, \textit{PKD} is a particular case of our model, but \textit{CKD} gives us more flexibility in terms of distilling from different teacher configurations.

\section{Experimental Study}\label{exp}
\begin{table*}[th]
\centering
\begin{tabular}{l m{35mm} c l l l l }
\toprule
\multicolumn{2}{c}{Models}& Pt$\rightarrow$En & Tr$\rightarrow$En & En$\rightarrow$De$_1$ & En$\rightarrow$De$_2$\\[0.5ex] 
\hline
\parbox[t]{1mm}{\multirow{4}{*}{\rotatebox[origin=c]{90}{\textit{Baselines}}}}&\textit{Teacher}& $43.69$ & $26.44$ &$18.57$ & $27.03$\\
&\textit{No-KD}&$42.12$ &$24.60$& $17.04$ & $16.09$\\
&\textit{Regular KD}& $42.26$ & $25.16$ &$17.66$ & $16.99$\\
&\textit{PKD}& $42.27$ & $26.88$ &$17.84$ & $21.06$\\\hline
\parbox[t]{1mm}{\multirow{4}{*}{\rotatebox[origin=c]{90}{\textit{CKD (Ours)}}}} &\textit{Regular Comb. (RC)}&$43.43$ & $26.75$&$18.29$ & $21.15$\\
&\textit{Overlap Comb. (OC)}& $\textbf{43.78}$ & $26.52$ & $18.44$ & $21.26$ \\
&\textit{Skip Comb. (SC)}& $43.17$ & $26.37$ & $17.81$ & $\textbf{21.47}$\\
&\textit{Cross Comb. (CC)}& $42.57$ & $\textbf{27.09}$ &$\textbf{18.60}$& $21.13$\\
\bottomrule
\end{tabular}
\caption{\label{t:1} BLEU score comparisons of different KD models. \textit{No-KD} is a model trained with no KD techniques using the the same architecture and dataset as students'.}
\end{table*}
Although our proposed model is a general KD technique and can be applied in different settings, we narrow down the scope of this paper to low-resource, NMT settings. The incentive idea behind our project was to train NMT models for small datasets, so we report experimental results accordingly. 

To evaluate \textit{CKD}, we trained multiple models to translate from English (En) into German (De), and from Portuguese (Pt) and Turkish (Tr) into English (En). For the Pt$|$Tr$\rightarrow$En directions we use the {\fontfamily{pcr}\selectfont IWSLT-2014} dataset, and the En$\rightarrow$De experiment uses the {\fontfamily{pcr}\selectfont WMT-2014} dataset.

In Pt$\rightarrow$En, we use the original split of datasets from IWSLT\footnote{\url{https://wit3.fbk.eu/}} with $167$K, $7590$, and $5388$ sentences for training, development, and test sets, respectively. For Tr$\rightarrow$En, the split is $142$K, $1958$, and $1982$ for training, development, and test sets. With this dataset selection our $\mathcal{T}$ models' results are comparable to publicly reported results.\footnote{\url{http://cs.jhu.edu/~kevinduh/a/multitarget-tedtalks/}} On these datasets, our teachers outperform all other existing models so we can ensure that we distill from reliable sources. 

For En$\rightarrow$De, the dataset is the same as the original Transformer's \citep{vaswani2017attention}, namely the training set includes $4.5$M sentences, {\fontfamily{pcr}\selectfont newstest2013} is used as the validation set and {\fontfamily{pcr}\selectfont newstest2014} is our test set with $3000$ and $3003$ sentences,  respectively. We selected this dataset to be comparable to a well-known baseline and make sure our training pipeline yields high-quality engines. 

We preprocess datasets with Sentence-Piece \citep{kudo2018sentencepiece}. For Pt$\rightarrow$En, we extracted a shared vocabulary set for both source and target sides with $32$K subwords. Both $\mathcal{S}$ and $\mathcal{T}$ are trained using the same training set. Tr$\rightarrow$En follows the same setting. For En$\rightarrow$De, we conduct two experiments. Since our focus in this paper is to work with low-resource settings, in En$\rightarrow$De$_1$, $\mathcal{S}$ and $\mathcal{T}$ are trained on a dataset of $200$K sentences randomly sampled from the main dataset ($4.5$M).\footnote{Our code and datasets: \url{https://github.com/yimeng0701/CKD_pytorch}} For this experiment the vocabulary set size is $15$K. In En$\rightarrow$De$_2$, we slightly changed the setting where we use the entire set of $4.5$M sentences to train $\mathcal{T}$ but $\mathcal{S}$ still uses the same $200$K dataset. In this scenario, we assumed that there already exists a high-quality teacher trained on a large dataset but we only have a small in-house dataset to train the student. For this experiment the vocabulary size is $37$K. 

Table \ref{t:1} summarizes our results for all experiments. Models are compared based on BLEU \citep{papineni2002bleu} scores computed using sacreBLEU \citep{post-2018-call}. As the table shows, our students outperform all other students trained with different KD techniques. Moreover, students in Pt$|$Tr$\rightarrow$En and En$\rightarrow$De$_1$ settings are even comparable to accurately-trained, deep teachers. All teachers are $12$-layer Transformers ($6$ for encoding and $6$ for decoding), whereas students only have $4$ layers ($2$ encoder layers and $2$ decoder layers). All settings in our experiments are identical to those of \citet{vaswani2017attention}, which means hyper-parameters whose values are not clearly declared in this paper use the same values as the original Transformer model. 

\textit{CKD} makes it possible to reduce the number of parameters in our students by $50\%$ and yet deliver the same high-quality translations. Accordingly, this enables us to run these translation engines on edge devices. Table \ref{t:par} shows the exact number of parameters for each model.
\begin{table}[h]
\begin{tabular}{l c c c c}
\toprule
& Pt$\rightarrow$En & Tr$\rightarrow$En & En$\rightarrow$De$_1$ & En$\rightarrow$De$_2$\\[0.5ex] 
\hline
$\mathcal{T}$&$61$M & $61$M &$52$M & $63$M\\
$\mathcal{S}$&$31$M &$31$M & $22$M & $34$M\\
\bottomrule
\end{tabular}
\caption{\label{t:par}The exact number of parameters for different models and settings.}
\end{table}

For results reported in Table \ref{t:1}, cross-model layer mappings between teacher and student layers are as follow: 
\begin{align*}
M_{SC}(1)&=\{1,2\} & M_{SC}(2)&=\{5,6\}\\
M_{CC}(1)&=\{1,3\} & M_{CC}(2)&=\{4,6\}\\
M_{RC}(1)&=\{1,2,3\} & M_{RC}(2)&=\{4,5,6\}\\
M_{OC}(1)&=\{1,2,3,4\} & M_{OC}(2)&=\{3,4,5,6\}
\end{align*}
We tried a simple (and somewhat arbitrary) configuration for layer connections and there is no systematic strategy behind it. However, better results can be achieved with better heuristics or through a search process. Moreover, as the mappings show there is no connection between student and teacher models' decoder layers. In our experiments, we noticed that \textit{any} KD technique applied to the decoder \textit{considerably} decreases performance, so we only use KD on the encoder side. More specifically, each student model has two decoder layers which only receive inputs from the same model's encoder layers and they are not connected to the teacher side.  

To train students with different KD techniques we use different loss functions. In $\mathcal{T}$ and \textit{No-KD} we only have a single loss function ($\mathcal{L}$) as described in the original Transformer model \citep{vaswani2017attention}. For models trained with \textit{RKD}, an additional loss is involved to match teacher and student predictions ($\mathcal{L}_{KD}$). The final loss in this case is an interpolation of the aforementioned losses: $\big((\beta \times \mathcal{L}) + (\eta \times \mathcal{L}_{KD})\big)$. In our experiments, $\beta = (1-\eta)$ where $\eta=0.1$ is obtained through a search process over the set $\{0.1, 0.3, 0.5, 0.7, 0.9\}$. 

For students trained using \textit{PKD} and \textit{CKD}, a third loss is also used in addition to $\mathcal{L}$ and $\mathcal{L}_{KD}$. Similar to other losses, the third one is also multiplied by a weight value ($\lambda$) to incorporate its impact into the training process. In this new setting, $\beta = (1 - \eta - \lambda$), $\eta = 0.1$, and $\lambda = 0.7$. The high value of $\lambda$ compared to other weights shows the importance of intermediate KD for deep models. All these values are learned through an empirical study in order to minimize the final loss of translation engines. 

\subsection{How Powerful is \textit{CKD}?}
In order to study the behaviour of \textit{CKD}, we designed multiple, small experiments in addition to those reported in Table \ref{t:1}. \textit{PKD} proposes a solution to define a loss between internal components of teacher and student models. The original model implemented this idea for intermediate layers. In one of our experiments we extended \textit{PKD} by adding an extra loss function for \textit{self-attention} components. Therefore, this new extension compares final outputs of student and teacher models as well as their intermediate layers and self-attention parameters. In this experiment, BLEU for Pt$\rightarrow$En increased from $42.27$ to $43.28$, but our model is still superior with the BLEU score $\textbf{43.78}$. For this setting, \textit{CKD} outperforms even a very complicated variation of \textit{PKD} that could be an indication of our model's capacity in training high-quality students. For Tr$\rightarrow$En and En$\rightarrow$De$_1$ we also observed slight improvements by matching teacher and student self-attention components but results were not statistically significant and \textit{CKD} was still better. 

We also studied how \textit{CKD} behaves in large experimental settings, for which we used En$\rightarrow$De and En$\rightarrow$French (Fr) datastes with $4.5$M and $36$M training samples, respectively, and trained $12$-layer teachers and $4$-layer students. For this experiment, we used the same settings, and test and development sets suggested in \citet{vaswani2017attention}. Table \ref{t:2} summarizes our results.\footnote{Performing a comprehensive study for large datasets was not our priority but we recently started related investigations and will report more inclusive results soon.} 
\begin{table}[h]
\centering
\begin{tabular}{m{10mm} m{7mm} m{12mm} m{7mm} m{7mm} m{7mm} }
\toprule
&$\mathcal{T}$ & \textit{No-KD} & \textit{PKD} & \textit{RC} & \textit{OC} \\\hline 
En$\rightarrow$Fr & $38.41$ & $35.45$ & $34.97$ & $36.10$ & $35.85$ \\
En$\rightarrow$De & $27.03$ & $24.31$ & $23.38$ & $24.14$ & $23.97$\\
\bottomrule
\end{tabular}
\caption{\label{t:2} BLEU scores of different KD models for large datasets.}
\end{table}

As the table shows, \textit{CKD} is better than \textit{PKD} in large experimental settings too. However, in order to have a better understanding of the large-dataset scenario we need to explore more configurations. We emphasize that for this paper our focus was to work with small students and datasets. 

\section{Conclusion and Future Work}\label{future}
In this paper, we proposed a novel model to distill from intermediate layers as well as final predictions. Moreover, we addressed the \textit{skip} problem of \textit{PKD}. We applied our technique in NMT and showed its potential in training high-quality and compact models. In our future work, \textit{i}) we are interested in distilling from deep NMT models into extremely small students with \textit{CKD}, in the hope of achieving the same results of large models with much smaller counterparts. \textit{ii}) We also try to improve the combination module and find a better alternative than concatenation. \textit{iii}) Finally, we plan to evaluate \textit{CKD} in other tasks such as language modeling. 

\section*{Acknowledgement}
We would like to thank our anonymous reviewers for their valuable feedback. 
\bibliography{emnlp2020}
\bibliographystyle{acl_natbib}

% \appendix

% \section{Appendices}
% \label{sec:appendix}

% \section{Supplemental Material}
% \label{sec:supplemental}

\end{document}